\documentclass[sigconf,9pt]{acmart}

\usepackage{booktabs}      
\usepackage{multirow}      
\usepackage{graphicx}

\usepackage{subcaption}
\usepackage[ruled,vlined]{algorithm2e}

\begin{document}
\newcommand{\system}{EPnG}

\acmYear{2026}\copyrightyear{2026}
\setcopyright{cc}
\setcctype[4.0]{by}
\acmConference[MobiSys Workshop '26]{The 24th Annual International Conference on Mobile Systems, Applications and Services}{June 21--25, 2026}{Cambridge, United Kingdom}
\acmBooktitle{The 24th Annual International Conference on Mobile Systems, Applications and Services (MobiSys Workshop '26), June 21--25, 2026, Cambridge, United Kingdom}
\acmDOI{10.1145/3812836.3814761}
\acmISBN{979-8-4007-2712-2/26/06}



\title{\system{}: Adaptive Expert Prune-and-Grow for Parameter-Efficient MoE Fine-tuning}

\author{Ahin Lee}
\affiliation{
  \institution{Ulsan National Institute of Science \& Technology (UNIST)}
  \city{Ulsan}
  \country{Republic of Korea}
}
\email{ahin@unist.ac.kr}

\author{Sehyun Yun}
\affiliation{
  \institution{Ulsan National Institute of Science \& Technology (UNIST)}
  \city{Ulsan}
  \country{Republic of Korea}
}
\email{nawhji@unist.ac.kr}

\author{Taesik Gong}
\affiliation{
  \institution{Ulsan National Institute of Science \& Technology (UNIST)}
  \city{Ulsan}
  \country{Republic of Korea}
}
\email{taesik.gong@unist.ac.kr}

\begin{abstract}

Mixture-of-Experts (MoE) models scale efficiently but remain costly to adapt due to redundant experts and uniform parameter allocation. Existing parameter-efficient fine-tuning (PEFT) methods such as LoRA ignore MoE routing dynamics, leading to suboptimal resource use. We propose EPnG, an adaptive prune-and-grow framework that reallocates LoRA capacity based on expert importance derived from router gate probabilities. EPnG prunes under-utilized experts and expands high-importance experts via rank growth with orthogonal initialization, while maintaining a fixed parameter budget.
Across OLMoE and Qwen1.5-MoE, EPnG consistently outperforms LoRA under the same budget and achieves performance comparable to full fine-tuning while updating only 0.55\%–0.72\% of parameters (up to 140x-180x fewer). These results demonstrate that aligning PEFT with MoE routing yields a more effective and scalable fine-tuning strategy.
\end{abstract}
\maketitle
\section{Introduction}

Mixture-of-Experts (MoE) architectures enable scalable large language models (LLMs) by activating only a subset of experts per token \citep{shazeer2017outrageously,fedus2022switch,Grattafiori2024Llama3Herd,achiam2023gpt,liu2024deepseek}. However, this efficiency does not extend to fine-tuning, where updating all experts remains prohibitively expensive \citep{rajbhandari2022deepspeed,kim2021scalable,aminabadi2022deepspeed}.
These challenges are further compounded by expert imbalance: some experts 
are consistently over-utilized while others are rarely 
activated~\citep{lepikhin2021gshard,fedus2022switch}, making uniform 
parameter allocation particularly wasteful during adaptation.

Parameter-efficient fine-tuning (PEFT) methods such as LoRA ~\citep{hu2022lora,lester2021power,li2021prefix} reduce training cost 
but ignore MoE routing dynamics, applying updates uniformly across experts. 
In dense models, every parameter receives consistent gradient updates across 
tokens. In contrast, MoE layers activate only a small subset of experts per 
token, resulting in sparse and uneven parameter updates. This discrepancy 
means that uniform LoRA allocation leaves critical experts under-adapted 
while wasting capacity on rarely selected 
ones~\citep{lepikhin2021gshard,fedus2022switch,zhang2023adalora,merchant2020happens}. 
Existing MoE-aware PEFT approaches either lack effective reallocation 
mechanisms or incur substantial parameter 
overhead~\citep{wang2024let,liu2024perft}.

We propose \textbf{\system{}} (Expert Prune-and-Grow LoRA), an adaptive PEFT framework for efficient MoE fine-tuning. \system{} estimates expert importance from router statistics, prunes low-importance experts, and reallocates the released budget to high-importance experts by expanding their LoRA ranks with orthogonal initialization. This dynamic reallocation aligns parameter allocation with routing behavior and concentrates capacity on task-relevant experts.

Across OLMoE and Qwen1.5-MoE, \system{} consistently outperforms static LoRA under the same parameter budget and achieves performance comparable to full fine-tuning while updating over 140× fewer parameters. These results demonstrate that coupling PEFT with MoE routing yields a more effective and scalable fine-tuning strategy.

\begin{figure}[t]
    \centering
    \includegraphics[width=0.98\linewidth]{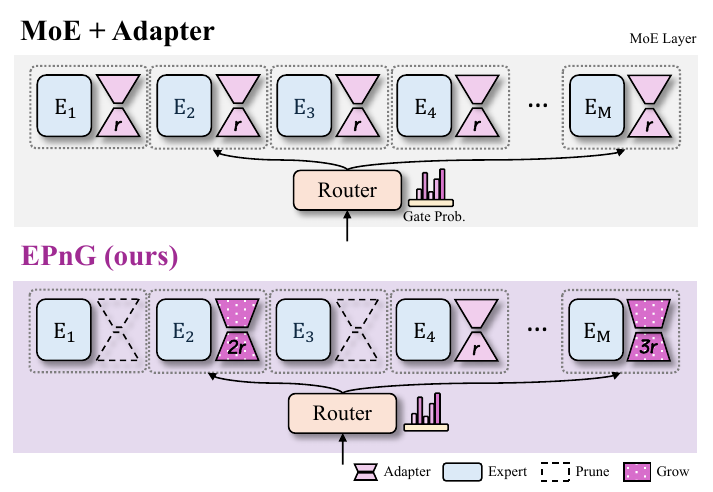}
    \caption{Standard MoE with uniform LoRA allocation (top) vs. EPnG (bottom). EPnG prunes low-importance experts and increases the ranks of frequently selected ones, yielding an importance-aware allocation under a fixed budget.}
    \Description{A two-part comparison figure. The top panel shows a standard MoE with uniform LoRA allocation across experts. The bottom panel shows EPnG, where less important experts are pruned and more frequently selected experts receive higher-rank adaptation under the same parameter budget.}
    \label{fig:epng}
\end{figure}
\section{Background}

\paragraph{Mixture-of-Experts.}
Mixture-of-Experts (MoE) architectures scale large language models (LLMs) via conditional computation \citep{Grattafiori2024Llama3Herd, achiam2023gpt, liu2024deepseek}. Given an input token representation $x \in \mathbb{R}^d$, a router selects the top-$k$ experts from $\mathcal{E} = \{E_1, \dots, E_M\}$. The router produces logits
$h(x) = W_\text{router} x \in \mathbb{R}^M$, which are normalized via softmax to obtain gate probabilities:
\begin{equation}
p_i(x) = \frac{e^{h(x)i}}{\sum_{j=1}^M e^{h(x)j}}, \quad i = 1, \dots, M.
\end{equation}
The selected expert set is
$\mathcal{S}(x) = \operatorname{Top-k}(p(x)), \ |\mathcal{S}(x)| = k,$
and the MoE output is
\begin{equation}
\mathrm{MoE}(x) = \sum_{i \in \mathcal{S}(x)} p_i(x) E_i(x)
\end{equation}

While this sparse activation enables efficient scaling, expert utilization is highly imbalanced in practice \citep{shazeer2017outrageously,chi2022representation}, motivating the need to quantify expert importance via routing signals.

\paragraph{Low-Rank Adaptation.}
Low-Rank Adaptation (LoRA) \citep{hu2022lora} enables parameter-efficient fine-tuning by injecting low-rank updates into pretrained weights. For $W \in \mathbb{R}^{d \times d}$:
\begin{equation}
W' = W + \Delta W, \quad \Delta W = AB,
\end{equation}
where $A \in \mathbb{R}^{d \times r}$, $B \in \mathbb{R}^{r \times d}$, and $r \ll d$. This reduces trainable parameters from $d^2$ to $2dr$.

\paragraph{Challenges in MoE Fine-Tuning.}

In dense models, parameters are uniformly updated across tokens, whereas MoE activates only a subset of experts, leading to sparse and uneven updates. Applying LoRA uniformly across experts thus results in inefficient parameter allocation, motivating MoE-aware adaptation strategies.
\begin{figure*}
    \centering
    \includegraphics[width=1.0\linewidth]{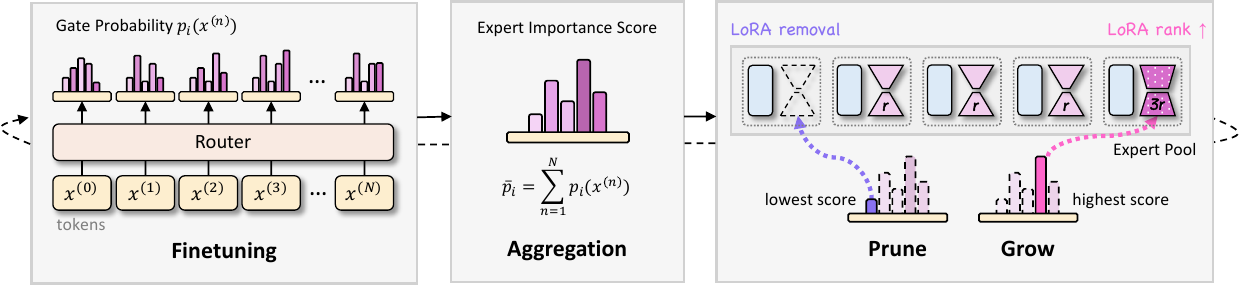}
    \caption{Illustration of \system{}'s end-to-end training procedure.}
    \label{fig:prune_and_grow}
\Description{Illustration of \system{}'s end-to-end training procedure.}

\end{figure*}

\section{Methodology}

We propose \textbf{\system{}} (Expert Prune-and-Grow), an adaptive PEFT framework for efficient MoE fine-tuning. As illustrated in Figure~\ref{fig:prune_and_grow}, \system{} collects router gate probabilities during training, aggregates them into expert importance scores, prunes low-importance experts, and reallocates the released budget by expanding the LoRA ranks of high-importance experts. This prune-and-grow loop dynamically concentrates capacity on task-relevant experts under a fixed parameter budget.

\subsection{Expert Importance Score Aggregation}

We estimate expert importance from router gate probabilities collected during fine-tuning (Figure~\ref{fig:prune_and_grow}, left). For each token $x^{(n)}$, the router outputs a distribution over experts, $\{p_i(x^{(n)})\}_{i=1}^M$. We aggregate these probabilities across tokens by averaging them (Figure~\ref{fig:prune_and_grow}, middle):
\begin{equation}
\bar{p}_i = \frac{1}{N} \sum_{n=1}^N p_i(x^{(n)}),
\end{equation}
where $N$ is the total number of processed tokens. We refer to $\bar{p}_i$ as the \textit{Expert Importance Score} of expert $E_i$. This score is the basis for pruning and rank reallocation.

\subsection{Pruning Low-Importance Experts}
\label{subsec:prune}

Using the computed importance scores, we identify underutilized experts for pruning and reallocate the released budget in the growing stage (Section~\ref{subsec:grow}).

\paragraph{Expert Selection for Pruning.}
We define the pruning threshold $\tau_p$ such that the lowest $\alpha$ fraction of expert importance scores $\{\bar{p}_i^{(l)}\}$ fall below it, where $\alpha \in (0,1)$ is the pruning ratio. The pruned set is
\begin{equation}
\mathcal{P} = \{ E_i^{(l)} \;|\; \bar{p}_i^{(l)} \leq \tau_p,\; l \in \{1,\dots,L\},\; i \in \{1,\dots,M\} \}.
\end{equation}
Here, $E_i^{(l)}$ denotes expert $i$ in layer $l$, $\bar{p}_i^{(l)}$ its importance score, $L$ the number of MoE layers, and $M$ the number of experts per layer. Repeating this thresholding for $t$ cycles yields a cumulative pruned fraction of $1 - (1-\alpha)^t$.

\paragraph{Pruning Operation.}
For each selected expert $E_i^{(l)} \in \mathcal{P}$, we remove its LoRA parameters:
\begin{equation}
    \Delta W_i^{(l)} = 0, \quad \forall E_i^{(l)} \in \mathcal{P}.
\end{equation}
This removes the expert's LoRA contribution and optimizer state, freeing parameter budget for growth.

\subsection{Growing High-Importance Experts}
\label{subsec:grow}

We reallocate the budget released by pruning to highly utilized experts by expanding their LoRA ranks. We focus on the budget-neutral setting, where growth matches the budget released by pruning.

\paragraph{Expert Selection for Growth.}
We define the growth threshold $\tau_g$ such that the top $\beta$ fraction of importance scores $\{\bar{p}_i^{(l)}\}$ exceed it, where $\beta \in (0,1)$ controls the growth ratio. The selected set is
\begin{equation}
\mathcal{G} = \{ E_i^{(l)} \;|\; \bar{p}_i^{(l)} \geq \tau_g,\; l \in \{1,\dots,L\},\; i \in \{1,\dots,M\} \}.
\end{equation}
Repeated application for $t$ cycles yields a cumulative grown fraction of $1 - (1-\beta)^t$.

\paragraph{Rank Expansion.}
For each expert $E_i^{(l)} \in \mathcal{G}$, we expand its LoRA rank from $r_i^{(l)}$ to $r_i^{(l)} + \Delta r$ by appending new low-rank components. If the original update is $\Delta W_i^{(l)} = A_i^{(l)} B_i^{(l)}$, the expanded update becomes:
\begin{equation}
    \Delta W_i^{(l)\prime} = 
    [A_i^{(l)} \ \ A_i^{\text{new},(l)}] 
    \begin{bmatrix} B_i^{(l)} \\ B_i^{\text{new},(l)} \end{bmatrix}.
\end{equation}

\paragraph{Initialization and Orthogonalization.}
To ensure stable and non-redundant expansion, $A_i^{\text{new},(l)}$ is initialized with Kaiming initialization~\citep{he2015delving}, while $B_i^{\text{new},(l)}$ is initialized to zero, following LoRA~\citep{hu2022lora}. We further orthogonalize the new directions against the existing subspace:
\begin{equation}
    A_i^{\text{new},(l)} \leftarrow (I - Q_i^{(l)} {Q_i^{(l)}}^\top) A_i^{\text{new},(l)},
\end{equation}
where $Q_i^{(l)} = \text{orth}(A_i^{(l)})$ is an orthonormal basis spanning the columns of $A_i^{(l)}$.

\subsection{Prune-and-Grow Adaptation Loop}

\paragraph{Warm-up for Stable Importance Estimation.}
Because expert importance evolves during fine-tuning, we first use a warm-up stage of $T_w$ steps, during which gate statistics are collected but no pruning or growth is applied.

\paragraph{Adaptive Parameter Reallocation.}
After warm-up, the prune-and-grow procedure is triggered every $T_p$ steps. At each interval, we compute $\{\bar{p}_i^{(l)}\}$, prune low-importance experts (Section~\ref{subsec:prune}), and expand high-importance experts (Section~\ref{subsec:grow}). Under the constraint $\alpha \geq \beta$ and assuming uniform rank expansion per expert, pruning releases equal or more parameters than growth consumes, ensuring that the total number of trainable parameters does not exceed the initial budget.

\paragraph{Summary.}
The prune-and-grow loop thus turns finetuning into a dynamic, budget-aware adaptation process. 
By reallocating LoRA parameters according to router-derived importance signals, 
it promotes (i) consistent utilization of the limited parameter budget, 
(ii) reinforcement of task-relevant experts, and 
(iii) suppression of redundant adaptations. 
Algorithm~\ref{alg:\system{}} provides the pseudocode for the full training loop, and the overall workflow is illustrated in Figure~\ref{fig:prune_and_grow}.

\begin{algorithm}[ht]
\caption{Expert Prune-and-Grow (\system{}) Training}
\label{alg:\system{}}
\KwIn{
    MoE-LLM with initial LoRA rank $r_0$, dataset $D$, warm-up steps $T_w$, prune interval $T_p$, pruning and growth ratios $\alpha$, $\beta$, decay factor $\lambda$
}
Initialize gate statistics $\bar{p}_i^{(l)} \leftarrow 0$ for all experts \\
\For{$step = 1$ \KwTo $max\_steps$}{
    Sample batch $x \sim D$ and compute loss \\
    Update gate probabilities $\bar{p}_i^{(l)}$ \\
    Optimize model parameters via backpropagation \\

    \If{$step \geq T_w$ \textbf{and} $step \bmod T_p = 0$}{
        Compute pruning threshold $\tau_p$ and growth threshold $\tau_g$ from $\{\bar{p}_i^{(l)}\}$ \\
        Select experts to prune: 
        $\mathcal{P} = \{ E_i^{(l)} \mid \bar{p}_i^{(l)} \leq \tau_p \}$ \\
        Select experts to grow: 
        $\mathcal{G} = \{ E_i^{(l)} \mid \bar{p}_i^{(l)} \geq \tau_g \}$ \\

        \ForEach{$E_i^{(l)} \in \mathcal{P}$}{
            Remove LoRA parameters: $\Delta W_i^{(l)} \leftarrow 0$
        }

        \ForEach{$E_i^{(l)} \in \mathcal{G}$}{
            Expand LoRA rank: $r_i^{(l)} \leftarrow r_i^{(l)} + \Delta r$ \\
            Initialize $A_i^{\text{new},(l)}$ with Kaiming init, $B_i^{\text{new},(l)} \leftarrow 0$ \\
            Orthogonalize: 
            $A_i^{\text{new},(l)} \leftarrow (I - Q_i^{(l)} {Q_i^{(l)}}^\top) A_i^{\text{new},(l)}$
        }

        Decay gate statistics: 
        $\bar{p}_i^{(l)} \leftarrow \lambda \bar{p}_i^{(l)}$
    }
}
\end{algorithm}

\section{Experiments}
\label{sec:experiments}

\subsection{Experimental Setup}
\label{sec:exp-setup}

\paragraph{Datasets and evaluation.}
We evaluate EPnG across three domains: mathematical reasoning, code generation, and personalization.

\textbf{Mathematical reasoning.}
We fine-tune on MetaMathQA~\citep{yu2024metamath}, a bootstrapped dataset of diverse math questions, and evaluate on GSM8K~\citep{cobbe2021training}, which consists of grade-school arithmetic word problems, and MATH~\citep{hendrycks2021measuring}, a large-scale benchmark covering competition-level mathematics.
Performance is measured by Exact Match (EM), i.e., the proportion of predictions that exactly match the ground-truth answer.

\textbf{Code generation.}
We fine-tune on Code Alpaca~\citep{luo2023wizardcoder}, an instruction-following dataset for programming tasks, and evaluate on HumanEval~\citep{chen2021evaluating} and MBPP~\citep{austin2021program}, two widely adopted benchmarks for code synthesis.
We report pass@10~\citep{chen2021evaluating}, which estimates the probability of generating at least one correct solution within ten independently sampled candidates.

\textbf{Personalization.}
We evaluate on PrefEval~\citep{zhao2025llms}, a benchmark designed to measure how well models infer and follow user-specific preferences in conversation.
Model outputs are assessed via pairwise comparison using GPT-4o-mini~\citep{achiam2023gpt} as an automatic evaluator, following the official preference following accuracy metric.

To ensure fair and reproducible comparison, all decoding and evaluation settings are fixed across methods.
For generation tasks, we use temperature $=0.8$, top-$p=0.95$, and a maximum generation length of 512 tokens.
For pass@10 evaluation, we sample 10 independent outputs per problem.
For PrefEval, we use identical prompts and random seeds across all methods.

\paragraph{Models and baselines.}
We conduct experiments on two public MoE backbones: allenai/OLMoE-1B-7B-0125~\citep{niklas2025olmoe} and Qwen/Qwen1.5-MoE-A2.7B~\citep{qwen15moe2024}. We compare four adaptation strategies: full fine-tuning (FFT), static LoRA, ESFT~\citep{wang2024let}, and EPnG. FFT updates all model parameters and serves as an upper-bound reference. Static LoRA applies a fixed rank to all experts. ESFT is a recent MoE-specific PEFT baseline, and EPnG adaptively reallocates LoRA capacity using expert importance scores.

\paragraph{Implementation details.}
LoRA adapters are applied to the \texttt{up\_proj} and \texttt{gate\_proj} matrices. Unless otherwise stated, we use pruning ratio $\alpha=0.2$, growth ratio $\beta=0.2$, warm-up steps $T_w=100$, update interval $T_p=50$, and exponential decay factor $\lambda=0.2$ for gate statistics. We follow the standard evaluation protocol for each benchmark: EM for GSM8K and MATH, pass@10 for HumanEval and MBPP, and pairwise preference evaluation on PrefEval using GPT-4o-mini. 

\begin{table}[t]
\centering
\caption{Results on OLMoE. Metrics: Exact Match (\%) for GSM8K/MATH, pass@10 (\%) for MBPP/HumanEval, and accuracy (\%) for PrefEval. Trainable parameters are reported as the average \% of the full model.}
\resizebox{0.48\textwidth}{!}{%
\begin{tabular}{l c cc cc c c}
\toprule
\multirow{2}{*}{Method} & \multirow{2}{*}{Params. (\%)} & \multicolumn{2}{c}{\textbf{Math (EM)}} & \multicolumn{2}{c}{\textbf{Code (pass@10)}} & \textbf{Personalization} & \multirow{2}{*}{Avg.} \\
\cmidrule(lr){3-4}\cmidrule(lr){5-6}
 &  & GSM8K & MATH & MBPP & HumanEval & PrefEval &  \\
\midrule
FFT           & 100   & 64.97 & 24.68 & 43.60 & 25.00 & 34.44 & 38.54 \\
\midrule
ESFT          & 5.08  & 65.13 & 22.49 & 42.20 & 24.39 & 36.11 & 38.06 \\
LoRA (static) & \textbf{0.72}  & 64.44 & 21.90 & 39.60 & 22.56 & 35.00 & 36.70 \\
Ours (EPnG)   & \textbf{0.72}  & 66.26 & 21.40 & 42.40 & 23.78 & 38.33 & \textbf{38.43} \\
\bottomrule
\end{tabular}%
}
\label{tab:olmoe_results_grouped}
\end{table}

\begin{table}[t]
\centering
\caption{Results on Qwen1.5-MoE. Trainable parameters are reported as the average \% of the full model.}
\resizebox{0.48\textwidth}{!}{%
\begin{tabular}{l c cc cc c}
\toprule
\multirow{2}{*}{Method} & \multirow{2}{*}{Params. (\%)} & \multicolumn{2}{c}{\textbf{Math (EM)}} & \multicolumn{2}{c}{\textbf{Code (pass@10)}} & \multirow{2}{*}{Avg.} \\
\cmidrule(lr){3-4}\cmidrule(lr){5-6}
 &  & GSM8K & MATH & MBPP & HumanEval &  \\
\midrule
ESFT          & 15.91 & 64.14 & 34.54 & 60.40 & 76.83 & \textbf{58.98} \\
LoRA (static) & \textbf{0.55}  & 62.93 & 34.96 & 60.40 & 70.12 & 57.60 \\
Ours (EPnG)   & \textbf{0.55}  & 64.22 & 37.32 & 60.00 & 74.39 & \textbf{58.98} \\
\bottomrule
\end{tabular}%
}
\label{tab:qwen_results}
\end{table}

\subsection{Main Results}
\label{sec:main-results}

\paragraph{Results on OLMoE}
Table~\ref{tab:olmoe_results_grouped} shows that EPnG updates only $0.72\%$ of parameters, over $140\times$ fewer than FFT, while achieving comparable overall performance. In particular, EPnG outperforms FFT on GSM8K ($66.26$ vs.\ $64.97$) and improves over static LoRA under the same budget by +1.82 points on GSM8K and +1.73 points on average. This shows that reallocating LoRA capacity across experts is more effective than uniform allocation.

\paragraph{Results on Qwen1.5-MoE}
Table~\ref{tab:qwen_results} shows that EPnG remains effective on a larger MoE backbone. Although EPnG tunes only $0.55\%$ of parameters, compared with $15.91\%$ for ESFT, it matches ESFT in average performance ($58.98$). Under the same budget as static LoRA, EPnG improves MATH from $34.96$ to $37.32$ and HumanEval from $70.12$ to $74.39$, indicating that dynamic expert-wise reallocation scales favorably.

\paragraph{Overall comparison.}
Across both backbones, EPnG consistently improves or matches stronger baselines while tuning less than $1\%$ of parameters. On OLMoE, it approaches FFT with over $140\times$ fewer trainable parameters; on Qwen1.5-MoE, it matches ESFT with about $29\times$ fewer. Overall, these results demonstrate that routing-aware prune-and-grow provides a scalable and parameter-efficient alternative to static LoRA and costly full fine-tuning.

\section{Further Analysis}
\label{sec:analysis}

\begin{figure*}[t]
  \centering
    \begin{subfigure}[t]{0.32\textwidth}
      \centering
      \vspace{-8.8em}
      \begin{tabular}{lc}
        \toprule
        \textbf{Method} & \textbf{Acc.} \\
        \midrule
        LoRA          & 65.50 \\
        Grow only     & 64.97 \\
        Prune only    & 65.81 \\
        Prune \& Grow & \textbf{66.41} \\
        \bottomrule
      \end{tabular}
      \vspace{0.48cm}
      \caption{Ablation study.}
      \Description{Ablation study.}
      \label{fig:analysis_ablation}
    \end{subfigure}
    \hspace{-0.03\textwidth} 
    \begin{subfigure}[t]{0.3\textwidth}
      \centering
      \includegraphics[width=\linewidth]{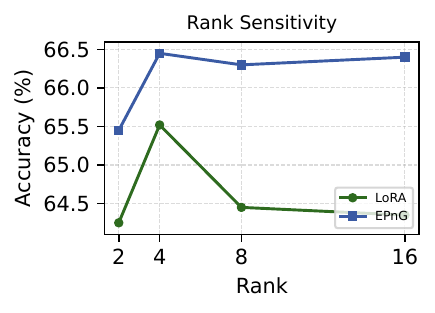}
      \caption{Rank sensitivity.}
      \label{fig:analysis_rank}
      \Description{Rank sensitivity.}
    \end{subfigure}
    \hfill
    \begin{subfigure}[t]{0.32\textwidth}
      \centering
      \includegraphics[width=\linewidth]{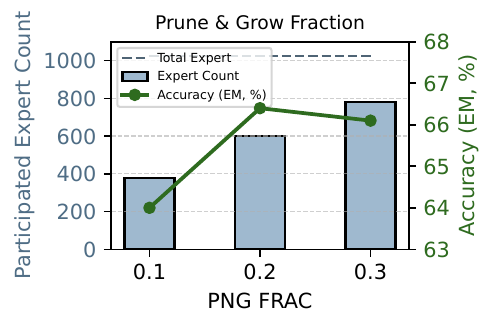}
      \caption{Effect of prune-and-grow fraction.}
      \Description{Effect of prune-and-grow fraction.}
      \label{fig:analysis_fraction}
    \end{subfigure}

  \caption{
  Analysis of EPnG. 
  (a) Ablation over pruning and growth.
  (b) Accuracy across different base LoRA ranks.
  (c) Effect of the prune-and-grow fraction on accuracy and expert usage.
  }
  \label{fig:analysis_three}
\end{figure*}

\begin{figure}[t]
  \centering
  \includegraphics[width=0.48\textwidth]{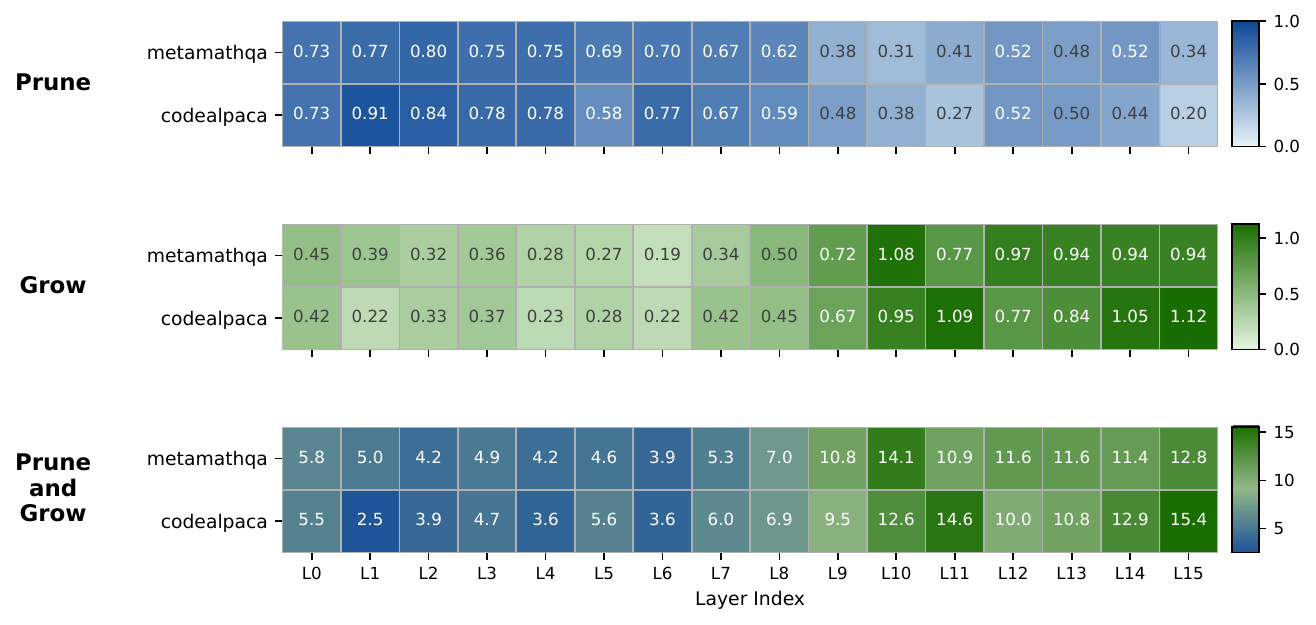}
  \caption{
  Layer-wise behavior of EPnG. 
  Top: pruning ratios across layers. 
  Middle: growth ratios across layers. 
  Bottom: resulting average rank distribution after repeated prune-and-grow updates. 
  EPnG prunes shallower layers more aggressively and reallocates capacity toward deeper layers.
  }
  \label{fig:analysis_layer}
\Description{Layer-wise behavior of EPnG. 
  Top: pruning ratios across layers. 
  Middle: growth ratios across layers. 
  Bottom: resulting average rank distribution after repeated prune-and-grow updates. 
  EPnG prunes shallower layers more aggressively and reallocates capacity toward deeper layers.}

\end{figure}

\begin{figure}[t]
    \centering
    \includegraphics[width=0.48\textwidth]{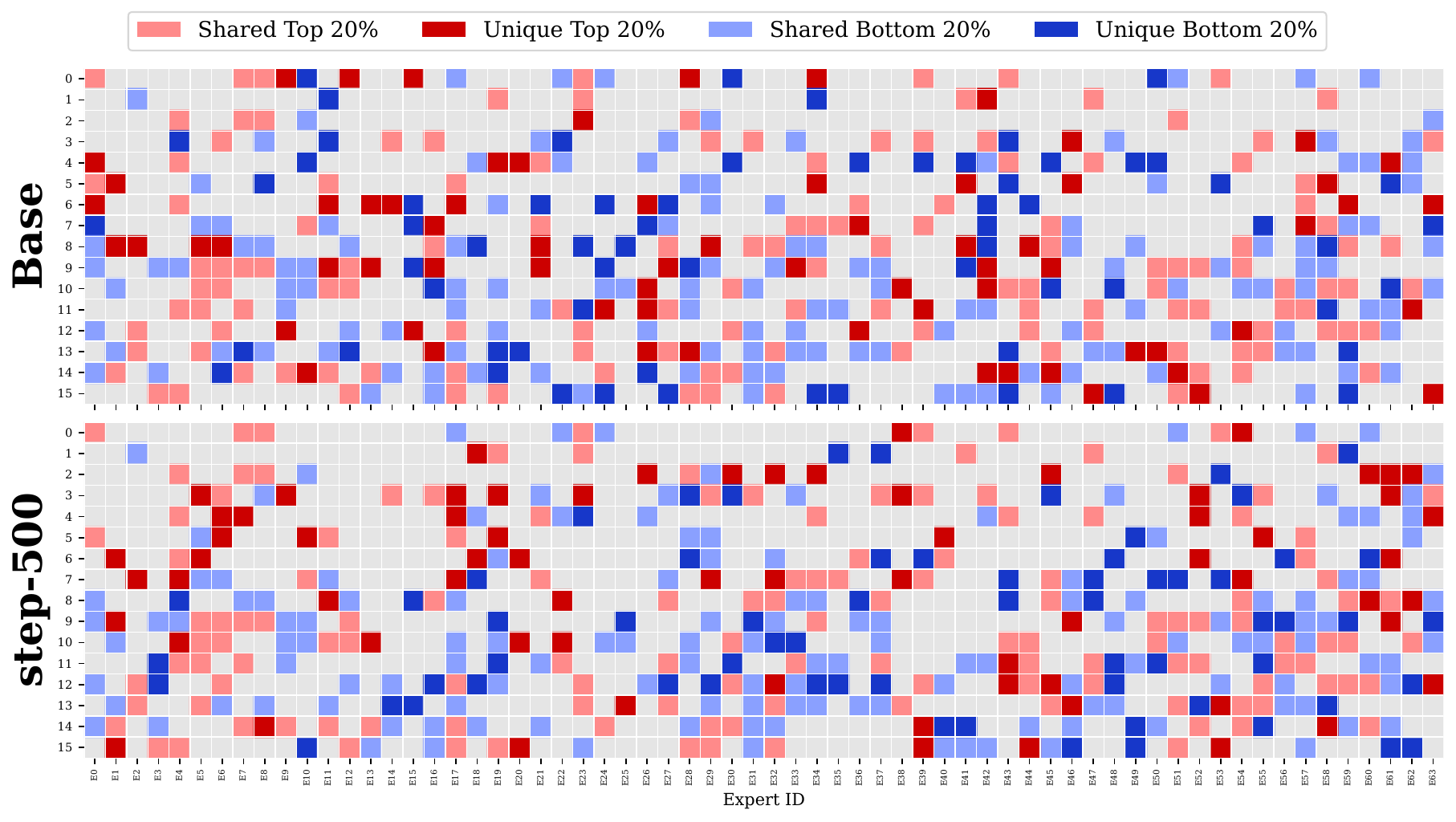}
    \caption{
    Expert importance before and after fine-tuning. 
    The top and bottom 20\% experts are highlighted, distinguishing experts whose importance is preserved (Shared) from those whose importance changes after fine-tuning (Unique). 
    Fine-tuning substantially reshapes the expert importance distribution.
    }
    \Description{Expert importance before and after fine-tuning. }
    \label{fig:expert_importance}
\end{figure}

We provide additional analyses to understand why EPnG improves performance under tight parameter budgets. Unless otherwise stated, all analyses are conducted on GSM8K with OLMoE. Figures~\ref{fig:analysis_three}--\ref{fig:expert_importance} and Tables~\ref{tab:gen}--\ref{tab:lora-placement-ablation} summarize the results.

\paragraph{Ablation study.}
We conduct an ablation study to examine the contributions of pruning and growing (Figure~\ref{fig:analysis_ablation}). Compared to the static LoRA baseline (65.50\%), applying either growing (64.97\%) or pruning (65.81\%) in isolation yields comparable or slightly improved performance. Notably, pruning proves more effective than growing, as it achieves better accuracy with fewer parameters than the baseline. In contrast, combining both operations leads to the best result (66.41\%), indicating that pruning and growing complement each other: pruning eliminates redundant components, while growing reinforces useful ones.

We speculate that pruning acts as an implicit regularizer, removing redundancy and encouraging the model to concentrate on essential representations. Growing, by contrast, may introduce noise without resolving redundancy, and newly added parameters can be underutilized early in training. Together, these effects explain why combining both operations outperforms either alone.


\paragraph{Rank sensitivity.}
Figure~\ref{fig:analysis_rank} shows that static LoRA is sensitive to the choice of base rank, with noticeable performance fluctuations across ranks. EPnG remains considerably more stable, suggesting that dynamic reallocation reduces the need for precise rank tuning.

\paragraph{Effect of the prune-and-grow fraction.}
Figure~\ref{fig:analysis_fraction} studies the prune-and-grow fraction. Performance improves as the fraction increases up to around 0.2, after which the gains saturate. This indicates that moderate reallocation provides the best trade-off between removing redundancy and preserving sufficient expert diversity.

\paragraph{Layer-wise behavior.}

Figure~\ref{fig:analysis_layer} visualizes how pruning and growing are distributed across layers, showing that shallower layers are pruned more heavily while deeper layers receive expanded ranks. This demonstrates that EPnG adaptively concentrates capacity toward more task-relevant layers rather than allocating it uniformly.

\paragraph{Expert importance dynamics.}
A key motivation for EPnG is that expert importance is not static during fine-tuning. To verify this, we compare expert importance scores before and after LoRA fine-tuning on 1,000 MetaMathQA examples using OLMoE.
Figure~\ref{fig:expert_importance} shows that the set of highly important experts changes substantially after fine-tuning: some experts remain consistently important, while others newly emerge or become less relevant. This confirms that static expert allocation is suboptimal and supports the need for repeated prune-and-grow updates during training.

\paragraph{Generalization after fine-tuning.}
To test whether EPnG preserves the backbone's general capability, we evaluate fine-tuned models on ARC-C, ARC-E, and BoolQ without task-specific adaptation. Table~\ref{tab:gen} shows that EPnG preserves performance close to the base model, whereas FFT degrades more noticeably. This suggests that EPnG improves task adaptation without substantially compromising general capabilities.

\begin{table}[t]
\centering
\caption{General evaluation after fine-tuning. EPnG largely preserves the general capability of the backbone.}
\begin{tabular}{lccc}
\toprule
Model & ARC-C & ARC-E & BoolQ \\
\midrule
Base & 50.17 & 69.30 & 66.54 \\
FFT  & 48.16 & 63.51 & 65.75 \\
LoRA & 50.17 & 68.42 & 65.23 \\
EPnG & 50.17 & 69.12 & 65.75 \\
\bottomrule
\end{tabular}
\label{tab:gen}
\end{table}


\paragraph{Effect of LoRA placement.}
We also study whether adding LoRA to \texttt{down\_proj} provides additional gains beyond the default \texttt{up\_proj}+ \texttt{gate\_proj} setting. As shown in Table~\ref{tab:lora-placement-ablation}, although this increases the parameter count from 0.72\% to 1.09\%, EPnG consistently maintains strong performance, with improvements observed as the parameter budget increases.

\begin{table}[t]
\centering
\caption{Effect of LoRA placement in OLMoE. 
EPnG consistently performs well across different adapter placements, demonstrating robustness to module configuration.
}
\begin{tabular}{lcccc}
\toprule
Method & Params. (\%) & GSM8K & MATH & Avg. \\
\midrule
LoRA (up+gate) & 0.72 & 64.44 & 21.90 & 43.17 \\
LoRA (+down)   & 1.09 & 65.35 & 22.33 & 43.84 \\
\midrule
EPnG (up+gate) & 0.72 & 66.26 & 21.40 & 43.83 \\
EPnG (+down)   & 1.09 & 66.19 & 23.50 & 44.85 \\
\bottomrule
\end{tabular}
\label{tab:lora-placement-ablation}
\end{table}

\paragraph{Summary.}
These analyses support the central design of EPnG. Pruning and growth are complementary, dynamic reallocation reduces rank sensitivity, moderate pruning works best, and the resulting capacity shifts concentrate on deeper, more task-relevant layers. Moreover, expert importance changes during fine-tuning, and general capability is preserved. Together, these findings explain why EPnG consistently outperforms static LoRA under the same parameter budget.



\section{Discussion}
A key limitation of EPnG is that pruning permanently removes LoRA parameters from under-utilized experts, making recovery impossible if a previously pruned expert becomes important in later training stages. To mitigate this risk, EPnG applies pruning gradually, targeting only the lowest-importance experts at each interval rather than aggressively removing a large fraction at once. This conservative strategy reduces the chance of prematurely discarding experts that may become task-relevant as training progresses. An alternative would be to retain pruned weights in memory and restore them when needed; however, this introduces non-trivial memory overhead and raises questions about whether retained parameters remain compatible with the evolved adapter subspace. We leave this direction to future work.

Beyond irreversibility, \system{} introduces additional hyperparameters compared to static LoRA, including the pruning ratio $\alpha$, growth ratio $\beta$, warm-up steps $T_w$, update interval $T_p$, and decay factor $\lambda$, which may impose non-trivial tuning costs when adapting to new models or tasks. However, our experiments show that a single fixed configuration generalizes well across different datasets and model backbones, suggesting that EPnG is not overly sensitive to hyperparameter choices in practice.

\section{Conclusion}
\label{sec:conclusion}
We introduced \textbf{\system{}} (Expert Prune-and-Grow), a 
parameter-efficient fine-tuning framework for Mixture-of-Experts 
(MoE) models that dynamically reallocates LoRA capacity based on 
router-derived expert importance. By pruning under-utilized experts 
and expanding high-importance ones under a fixed parameter budget, 
EPnG aligns adaptation with MoE routing dynamics.

Across OLMoE and Qwen1.5-MoE, \system{} achieves performance comparable 
to or better than full fine-tuning and ESFT while updating only about 
$1\%$ of parameters, and consistently outperforms static LoRA. 
Further analysis shows that EPnG stabilizes rank sensitivity and 
adaptively redistributes capacity toward deeper, more task-relevant 
layers, leading to more effective utilization of limited parameters.

Overall, \system{} demonstrates that coupling PEFT with dynamic expert 
allocation is a scalable and effective strategy for MoE fine-tuning 
under strict resource constraints. These results suggest that 
expert-wise parameter allocation should be treated as a dynamic 
optimization problem rather than a fixed design choice, an insight 
that opens broader opportunities for routing-aware adaptation in 
increasingly large and sparse language models.

\bibliographystyle{ACM-Reference-Format}
\bibliography{refs}

\end{document}